\documentclass{article}

\PassOptionsToPackage{}{natbib}

\usepackage[letterpaper]{geometry}
\usepackage[utf8]{inputenc} 
\usepackage[T1]{fontenc}    
\usepackage{hyperref}       
\usepackage{url}            
\usepackage{booktabs}       
\usepackage{amsfonts}       
\usepackage{nicefrac}       
\usepackage{microtype}      
\usepackage{amssymb}
\usepackage{amsmath}
\usepackage{amsthm} 
\usepackage{mathrsfs}
\usepackage{algorithm}
\usepackage{algpseudocode}
\usepackage{bm}
\usepackage{dsfont}
\usepackage[dvipsnames]{xcolor}
\usepackage{bbm}
\usepackage{natbib}
\usepackage{float}
\usepackage{xfrac}
\usepackage{todonotes}
\usepackage{graphicx}
\usepackage{stackengine}
\usepackage[para,online,flushleft]{threeparttable}
\usepackage{caption}
\usepackage{makecell}
\usepackage{booktabs}
\usepackage{array}
\newcolumntype{P}[1]{>{\centering\arraybackslash}p{#1}}
\newcolumntype{M}[1]{>{\centering\arraybackslash}m{#1}}

\newcommand{\R}{\mathcal{R}}

\newcommand{\inner}[1]{ \left\langle {#1} \right\rangle }
\newcommand{\inn}[1]{ \langle {#1} \rangle }
\newcommand{\Ber}{\operatorname{Bern}}
\newcommand{\p}{\prime}
\newcommand\norm[1]{\left\lVert#1\right\rVert}
\newcommand{\sik}{\sum_{i=1}^K}

\newcommand{\PP}{\mathbb{P}}
\newcommand{\Q}{\mathbb{Q}}

\newcommand{\E}{\varepsilon}

\newcommand{\basis}{{\boldsymbol{e}}}
\newcommand{\one}{\boldsymbol{1}}
\newcommand{\order}{\ensuremath{\mathcal{O}}}
\newcommand{\otil}{\ensuremath{\widetilde{\mathcal{O}}}}
\newcommand{\indicator}[1]{\mathbbm{1}_{#1}}
\newcommand{\Ex}{\mathbb{E}}
\newcommand{\KL}{\operatorname{KL}}

\newcommand{\cV}{\mathcal{V}}

\DeclareMathOperator*{\argmin}{argmin}

\newtheorem{theorem}{Theorem}
\newtheorem{lemma}[theorem]{Lemma}

\theoremstyle{definition}

\theoremstyle{remark}
\newenvironment{pf}[1][\proofname]{{\noindent\bfseries #1.}}{\qed}

\title{On Adaptivity in Information-constrained \\Online Learning}

\author{Siddharth Mitra\\
Chennai Mathematical Institute\\
\texttt{smitra@cmi.ac.in}
\and
Aditya Gopalan\\
Indian Institute of Science\\
\texttt{aditya@iisc.ac.in}
}

\date{}

\begin{document}

\maketitle

\begin{abstract}
We study how to adapt to smoothly-varying (`easy') environments in well-known online learning problems where acquiring information is expensive. For the problem of label efficient prediction,
  which is a budgeted version of prediction with expert
  advice, we present an online algorithm whose regret depends optimally on the number of labels allowed and $Q^*$ (the quadratic variation of the losses of the best action in hindsight), along with a parameter-free counterpart whose regret depends optimally on $Q$ (the quadratic variation of the losses of all the actions). These quantities can be significantly smaller than $T$ (the total time horizon), yielding an improvement over existing, variation-independent results for the problem. We then extend our
  analysis to handle label efficient prediction with bandit feedback, i.e., label efficient bandits. Our work builds upon
  the framework of optimistic online mirror descent, and leverages second
  order corrections along with a carefully designed hybrid
  regularizer that encodes the constrained information structure of the problem. We then consider revealing action partial monitoring games -- a version of label efficient prediction with additive information costs, which in general are known to lie in the
  \textit{hard} class of games having minimax regret of order 
  $T^{\nicefrac{2}{3}}$. We provide a strategy with an 
  $\mathcal{O}((Q^*T)^{\nicefrac{1}{3}})$ bound for revealing action
  games, along with one with a $\mathcal{O}((QT)^{\nicefrac{1}{3}})$ bound for
  the full class of hard partial monitoring games, both being strict
  improvements over current bounds.
\end{abstract}

\section{Introduction}
Online learning is a branch of machine learning that is concerned with the problem of dynamically optimizing utility (or loss) over time in the face of uncertainty, and gives valuable principles to reason about acting under uncertainty. The study of online learning has developed along two concrete
lines insofar as modeling the uncertain environment is concerned. On one hand, there is a rich body of work on learning in stochastic environments from an average-case point of view, such as i.i.d. multi-armed bandits (see for example the
survey of \citet{bubeck2012regret}), online learning in Markov
decision processes \citep{jaksch2010near,azar2017minimax}, stochastic
partial monitoring \citep{bartok2011minimax}, etc., which often yields performance guarantees that are strong but can closely depend on the stochastic models at hand. On the other hand, much work has been devoted to studying non-stochastic (or arbitrary or adversarial) models of environments from a worst-case point of view-- prediction with experts, bandits and partial monitoring problems to name a few \citep{lugosibook} -- which naturally yields rather pessimistic guarantees.

Recent efforts have focused on bridging this spectrum of modeling structure in online learning problems as arising from non-stochastic environments with loss function sequences exhibiting adequate temporal regularity. These include the derivation of first-order regret bounds or adapting to loss sequences with low loss of the best action \citep{allenberg2006hannan}, second-order bounds or adapting to loss sequences with low variation in prediction
with experts \citep{rakhlinsridharan,steinliang} and `benign' multi-armed
bandits \citep{hazankale,imppath, sparsity, weiluo}.

In this regard, this paper is an attempt to extend our understanding of adaptivity to low variation in several standard online learning problems where information comes at a  cost, namely label efficient prediction \citep{nicololep}, label efficient bandits and classes of partial monitoring problems
\citep{nbd1}.  In the process, we uncover new ways of using existing
online learning techniques within the Online Mirror Descent (OMD) family,
and partially make progress towards a program of studying the impact
of `easy' (i.e., slowly-varying) environments in information-constrained online learning and
partial monitoring problems. Our specific contributions are:
\begin{enumerate}
\item For the label efficient prediction game with expert advice, we give
a learning algorithm with a regret bound of $\order{\big(\sqrt{\nicefrac{Q^*T(\log K)}{n}}\big)}$ where
$Q^*$ is the quadratic variation of the best expert, $T$ is the time
horizon of the game, $K$ is the number of experts and $n$ is the bound on
label queries; the bound holds for all regimes except when $\nicefrac{n Q^*}{T} =
\otil(K^2)$. We follow this up with an algorithm with an
unconditional regret guarantee of $\order{(\sqrt{\nicefrac{QT(\log K)}{n}})}$ that holds for any label query budget $n$ and total quadratic variation $Q$. Our algorithms are based on the optimistic OMD
framework, but with new combinations of the negative entropy and log-barrier
regularization that are best suited to the label efficient game's information
structure. 
\item We generalize the results to label efficient bandits where one receives
bandit (i.e., for only the chosen expert) feedback at only up to $n$ chosen time instants, and obtain $\order{\big(\sqrt{\nicefrac{Q^*T K}{n}}\big)}$ regret. We also show that our upper bounds on regret for label efficient prediction and label efficient bandits are tight in their dependence on $Q$ and $n$ by demonstrating variation-dependent fundamental lower bounds on regret.
\item We show that adapting to low variation is also possible in the class of \textit{hard} partial monitoring games as per the taxonomy of partial monitoring problems by \citet{nbd1}, where we show an algorithm that achieves $\order{(\left(QT K \right)^{\nicefrac{1}{3}})}$ regret. To the best of our knowledge, this is the first algorithm exhibiting instance-dependent bounds for partial monitoring.
\end{enumerate}

\paragraph{Problem Setup and Notation}
A label efficient prediction game proceeds for $T$ rounds with $K \leq T$ arms or `experts'. In each round (time instant) $t$, the learner selects an arm $i_t \in [K] := {1,2,\dots,K}$. Simultaneously, the adversary chooses a loss vector $\ell_t \in [0,1]^K$ where $\ell_{t,i}$ is the loss of arm $i$ at time $t$. At each round, the learner can additionally choose to observe the full loss vector $\ell_t$, provided the number of times it has done so in the past has not exceeded a given positive integer $n \leq T$ that represents an information budget or constraint. We work in the {\em oblivious} adversarial setting where $\ell_t$ does not depend on the previous actions of the learner $i_1, i_2, \dots, i_{t-1}$; this is akin to the adversary fixing the (worst-possible) sequence of loss vectors in advance. The learner's goal is to minimize its expected regret defined as
\[ \max_{i^* \in [K]} \mathbb{E} \left[ \sum_{t=1}^T \ell_{t,i_t} - \sum_{t=1}^T \ell_{t,i^*} \right], \]
where the expectation is taken with respect to the learner's randomness. Given a convex function $\R$ over $\Omega$, we denote by $D_\R$ the Bregman divergence with respect to $\R$ defined as $D_\R (x,y)\triangleq \R(x)- \R(y)-\inner{\nabla \R(y), x-y} \: \forall x,y \in \Omega$. For any point $u \in \mathbb{R}^K$, we define the local norm at $x$ with respect to $\R$ as $\norm{u}_x = \sqrt{u^\top \nabla^2 \R(x) u}$ and the corresponding dual norm as $\norm{u}_{x,*} = \sqrt{u^\top \nabla^{-2} \R(x) u}$. We denote by $\epsilon$, the fraction of time we are allowed the full loss vector i.e.\ $\epsilon = \nicefrac{n}{T}$. The $\epsilon$ can be seen as a way to model the constraint on information defined by the problem. The quadratic variation for a loss vector sequence $l_1, \ldots, l_T$ is defined by $\smash{Q = \sum_{t=1}^T \norm{\ell_t - \mu_T}_2^2}$ with $\mu_s = \frac{1}{s}\sum_{t=1}^s \ell_s$. Additionally, the quadratic variation of the best arm(s) is $\smash{Q^* = \sum_{t=1}^T (\ell_{t,i^*} - \mu_{T,i^*})^2}$ where $\mu_{s,i} = \frac{1}{s}\sum_{t=1}^s \ell_{s,i}$ and $i^* = \argmin_{i \in [K]} \sum_{t=1}^T \ell_{t,i}$~.

\section{Key Ideas and Algorithms}\label{sec:alg}

\paragraph{Optimistic OMD} The underlying framework behind our algorithms is that of Online Mirror Descent (OMD)(see, for example \citet{hazanocobook}). The \textit{vanilla} update rule of (active) mirror descent can be written as $\smash{{x_t = \argmin_{x \in \Omega} \{ \inn{x, \tilde{\ell}_{t-1}}+D_{\R}(x,x_{t-1})\}} }$. On the other hand, our updates are:
\begin{align}
x_t &= \argmin_{x \in \Omega} \{ \inn{x, \epsilon m_t}+D_\R(x,x_t^\p) \} \label{eqn:update1}\\ 
x_{t+1}^\p &= \argmin_{x \in \Omega} \{ \inn{x, \epsilon\tilde{\ell}_t+a_t}+D_\R(x,x_t^\p) \} \label{eqn:update2}
\end{align}
\noindent
where $\epsilon = \nicefrac{n}{T}$~, $m_t$ corresponds to \textit{optimistic}\footnote{`Optimistic' is used to denote the fact that we would be best off if these estimates were exactly the upcoming loss. Indeed, if $m_t$ were $\ell_t$, it would be equivalent to 1-step lookahead, known to yield low regret.} estimates of the loss vectors (which we will also refer to as messages), and $a_t$ denotes a second order correction that we explicitly define later. Throughout the paper, $\tilde{\ell}_t$ is used to denote an (unbiased) estimate of $\ell_t$ that the learner constructs at time $t$. Optimistic OMD with second order corrections was first studied in \citet{weiluo}, whereas its Follow-the-Regularized-Leader (FTRL) counterpart was introduced earlier by \citet{steinliang}. Both of these approaches build upon the general optimistic OMD framework of \citet{rakhlinsridharan} and \citet{chiang}. We define our updates with \textit{scaled} losses and messages, where we reiterate that the scaling factor $\epsilon$ reflects the limitation on information. This scaling also impacts our second order corrections which are $\smash{\approx\eta\epsilon^2(\tilde{\ell}_t - m_t)^2}$. It is worthwhile to note that this is explicitly different from the $\smash{\eta\epsilon(\tilde{\ell}_t - m_t)^2}$ that one may expect in light of the analysis done in \citet{weiluo}, or the $\smash{\eta(\tilde{\ell}_t - m_t)^2}$ one would anticipate when following \citet{steinliang}. One may argue that our update rules are equivalent to dividing throughout by $\epsilon$, or put differently, by merging an $\epsilon$ into the step size, and this indeed true. However, the point we would like to emphasize is that no matter how one defines the updates, the second order correction $a_t$ can be seen to incorporate the problem dependent parameter $\epsilon$. This tuning of the second order correction based on $\epsilon$ is different from what one observes for the full information problem \citep{steinliang} or for bandits \citep{weiluo}. The second order corrections represent a further penalty on arms which are deviating from their respective messages, and these corrections are what enable us to furnish best arm dependent bounds. As usual, the arm we play is still sampled from the distribution $x_t$ given by equation \eqref{eqn:update1}.

\renewcommand{\arraystretch}{1.4}
\begin{table}[ht]
\centering
\begin{tabular}[t]{M{4.3cm}M{5cm}M{4.8cm}}
\toprule
\textsc{\textbf{Reference}} & \textbf{\textsc{Feedback}} & \shortstack{ \textbf{\textsc{Negentropy:Log-barrier}}\\ \textbf{\textsc{Regularizer Ratio Used}}} \\ \hline
     \citet{sparsity} & Bandit & $1:2\eta$ \\ \hline
     \citet{weiluo} & Bandit & $0:1$\\ \hline
     \citet{imppath} & Bandit & $\nicefrac{K}{\eta} : \nicefrac{1}{\eta} = K : 1$\\ \hline
     \citet{steinliang} & Full Information & $1:0$\\ \hline
     This work & Label Efficient-- Full Information & $\nicefrac{1}{\eta} : \nicefrac{1}{\eta K} = K:1$\\ \hline
     This work & Label Efficient-- Bandit Feedback & $0:1$ \\
\bottomrule
\end{tabular}
\caption{Choice of regularization (negative entropy vs. logarithmic barrier) in OMD for exploiting regularity}
  \label{table:regularization}
\end{table}

\paragraph{Challenges \& Our Choice of Regularization} We briefly discuss the challenges posed by label efficient prediction and how our choice of regularizer addresses these. When shifting away from the classical prediction with expert advice problem to any \textit{limited} feedback (i.e., over experts or arms) information structure, one usually works with importance-weighted estimates of the loss vectors constructed using the observed (limited) feedback (called inverse propensity weighting estimation). This is indeed the case with label efficient prediction, however, the probabilities in the denominator remain fixed at $\epsilon$, unlike in bandits where the $x_{t,i}$ in the denominator can be arbitrarily small.

Consequently, one may be led to believe that the standard negative entropic regularizer, as is typically used for full information  \citep{steinliang}, will suffice for the more general but related label efficient prediction. However, maintaining the $\smash{|\eta\tilde{\ell}_t| \leq 1}$ inequality which is standard in analyses similar to Exp3 imposes a strict bound of $\eta \leq \epsilon$. Since the low quadratic variation, on the other hand, would encourage one to set an aggressive learning rate $\eta$, this makes the applicability of the algorithm rather limited, and even then, with marginal gain. Put crisply, it is desirable that low quadratic variation should lead an algorithm to choose an aggressive learning rate, and negative entropy fails to maintain a `stability' property (in the sense of Lemma \ref{lem:conditionforstab}), key in obtaining OMD regret bounds, in such situations. The log-barrier regularizer, used by \citet{weiluo} for bandit feedback certainly guarantees this, however using log-barrier blindly translates to a $\sqrt{K}$ dependence on the number of arms $K$.

These challenges place label efficient prediction with slowly varying losses in a unique position, as one requires enough curvature to ensure stability, yet not let this added curvature significantly hinder exploration. Our solution is to use a hybrid regularizer, that is, a weighted sum of the negative entropic regularizer and the log-barrier regularizer:

\[ \R = \nicefrac{1}{\eta} \sik x_i \log x_i- \nicefrac{1}{\eta K} \sik \log x_i\]
This regularizer has been of recent interest due to the work of \citet{imppath}, and \citet{sparsity}, but the weights chosen for both components is highly application-specific and tends to reflect the nature of the problem. As reported above, we only require the log-barrier to guarantee stability, and therefore associate a small (roughly $\nicefrac{1}{K\eta}$) weight to it and a dominant mass of $\nicefrac{1}{\eta}$ to negative entropy. This fact is revealed in the analysis where we use the log-barrier component solely to satisfy Lemmas \ref{lem:stability1} and \ref{lem:conditionforstab}, following which it is essentially dispensed. The additional $\nicefrac{1}{K}$ factor part of the log-barrier weight is carefully chosen to exactly cancel the $K$ in the leading $K\log T$ term generated by the log-barrier component, and consequently, not have a $\sqrt{K}$ dependence on $K$ in the final regret bound.

\paragraph{Reservoir Sampling} When considering quadratic variation as a measure of adaptivity, a natural message to pass is the mean of the previous loss history, that is $\smash{m_t = \mu_{t-1} = \nicefrac{1}{t-1}\sum_{s=1}^{t-1}\ell_s}$. However, the constraint on information prohibits us from having the full history, and we therefore have to settle for some estimate of the mean. Reservoir sampling, first used in \citet{hazankale}, solves this very problem. Specifically, by allocating roughly $k(1 + \log T)$ rounds for reservoir sampling (where we choose $k$ to be $\log T$), reservoir sampling gives us estimates $\tilde{\mu}_t$  such that $\Ex[\tilde{\mu}_t] = \mu_t$, and $\text{Var}[\tilde{\mu}_t] = \nicefrac{Q}{kt}$. It does so by maintaining a carefully constructed reservoir of size $k$, the elements from which are then averaged to output the estimate of the mean. Our message $m_t$ at any time $t$ is the average of the vectors contained in the reservoir $S$. 

\subsection{Main Algorithm}\label{subs:mainalgo}

\begin{algorithm}[H]
\caption{\textsc{Adaptive Label Efficient Prediction}}\label{algo:main2}
\begin{algorithmic}[1]
\State \textbf{Input:} $\R = \nicefrac{1}{\eta} \sik x_i \log x_i- \nicefrac{1}{\eta K} \sik \log x_i$~, \\$\eta$~, $\epsilon$
\State \textbf{Initialize:} $x'_1 = \argmin_{x \in \Omega} \R(x)$
\For{$t = 1,2,\dots, T$}
	\State $d_t \sim \Ber(\epsilon)$
		\State $x_t = \argmin_{x \in \Omega} \left\{\inner{x,\epsilon m_t} + D_{\R} (x,x'_t)\right\}$ 
		\State Play $i_t \sim x_t$, and if $d_t=1$, observe $\ell_t$
	\State Construct $\tilde{\ell}_t = \frac{(\ell_t - m_t)}{\epsilon}\mathbbm{1}_{\{d_t=1\}} + m_t$
	\State Let $a_t = 6\eta\epsilon^2(\tilde{\ell}_t - m_t)^2$
	\State Update:
	\State $x'_{t+1} = \argmin_{x \in \Omega} \left\{\inner{x, \epsilon\tilde{\ell}_t + a_t} + D_{\R} (x,x'_t)\right\}$
\EndFor
\end{algorithmic}
\end{algorithm}

Algorithm \ref{algo:main2} builds upon the ideas presented above and as stated, is specifically for the label efficient prediction problem discussed thus far. The algorithms required for the extensions we provide in section \ref{sec:extensions} are based upon algorithm \ref{algo:main2}, although with a few minor differences. We specify those differences as and when required. Also, in the interest of brevity, we have excluded the explicit mentioning of the reservoir sampling steps. 
Before we proceed, we would like to cleanly state our choice of messages, loss estimates, and second order corrections used and this is done in Table \ref{table:directory}. Our messages, for all the sections will be $m_{t,i} = \tilde{\mu}_{t-1,i}$. Note that throughout the paper, the random variable $d_t = 1$ signifies that we ask for feedback at time $t$, and is $0$ otherwise. Additionally, note that we consider not exceeding the budget of $n$ in expectation, however, there is a standard reduction to get a high probability guarantee which can be found in \citet{lugosibook}.

\begin{table}[ht]
\centering
\begin{tabular}[t]{ccccc}
\toprule
\textbf{\textsc{Problem}} & \textbf{\textsc{Section}} & $\bm{\tilde{\ell}_{t,i} - m_{t,i}}$ &  $\bm{a_t}$ & \textbf{\textsc{Regret Bound}}\\ \midrule
	\shortstack{Label Efficient \\Prediction} & \ref{subs:mainalgo}, \ref{sec:analysis} &  $\frac{(\ell_{t,i} - m_{t,i})}{\epsilon}\mathbbm{1}_{\{d_t=1\}} $ & $6\eta\epsilon^2(\tilde{\ell}_t - m_t)^2$ & $\otil{\left( \sqrt{\nicefrac{Q^*T}{n}} \right)}$  \\ \midrule
	\shortstack{Label Efficient \\Bandits} & \ref{sec:leb} &  $\frac{(\ell_{t,i}-m_{t,i})}{\epsilon x_{t,i}}\mathbbm{1}_{\{d_t =1, i_t=i\}}$ & $6\eta\epsilon^2 x_{t,i}(\tilde{\ell}_t - m_t)^2$& $\otil{\left( \sqrt{\nicefrac{Q^*TK}{n}} \right)}$ \\ \midrule
	\shortstack{Revealing Action \\Games} & \ref{sec:revact} &  $\frac{(\ell_{t,i} - m_{t,i})\mathbbm{1}_{\{d_t=1\}}}{\alpha}\mathbbm{1}_{\{d_t=1\}}$ & $6\eta\alpha^2(\tilde{\ell}_t - m_t)^2$ & $\otil{\left( (Q^*T)^{1/3} \right)}$ \\ \midrule
	\shortstack{Hard \\Partial Monitoring} & \ref{sec:hardpm} &  $\frac{\left(\ell_{t,i} - m_{t,i}\right)}{x_{t,j}}\mathbbm{1}_{\{i_t = j\}}$ & 0 & $\order{\left( (QTK)^{1/3} \right)}$ \\ 
\bottomrule
\end{tabular}
\caption{Overview of loss estimates, second order corrections, and the corresponding upper bounds on regret}
  \label{table:directory}
\end{table}%

\section{Results and Analysis}\label{sec:analysis}

We now give a general regret result for the OMD updates \eqref{eqn:update1} and \eqref{eqn:update2}. It spells out the condition we must maintain to ultimately enable best arm dependent bounds while also demonstrating the price of limited information on regret, which is the additional $\nicefrac{1}{\epsilon}$ factor. The proofs for all results in this section appear in the supplementary material. 

\begin{lemma}\label{thm:generalthm}
For the update rules \eqref{eqn:update1} and \eqref{eqn:update2},
if:
\begin{align}
\inn{x_t-x^\p_{t+1}, \epsilon(\tilde{\ell}_t-m_t)+a_t} - \inn{x_t, a_t} \leq 0 \label{eqn:critical} 
\end{align}
then, for all $u\in \Omega$, we have:
\begin{align}
\inn{x_t-u, \tilde{\ell}_t}\leq \frac{1}{\epsilon} \left( D_\R(u,x_t^\p)-D_\R(u,x^\p_{t+1})+\inn{u,a_t} - P_t \right),
\end{align}
where $P_t \triangleq D_{\R}(x_{t+1}^\p, x_t)+D_{\R}(x_t, x_t^\p)\geq 0$
\end{lemma}
\noindent
Note that when $a_t = 0$ is employed in the updates \eqref{eqn:update1}-\eqref{eqn:update2}, i.e., no second order corrections, the first term in \eqref{eqn:critical} can directly be handled using H\"older's inequality (in some norm where $\R$ is strongly convex). Doing so allows us to cancel the unwanted $\smash{\|x_t-x^\p_{t+1}\|^2}$ term using the $\smash{ D_{\R}(x_{t+1}^\p, x_t)}$ term in $P_t$ (which follows by strong convexity) while retaining the crucial $\smash{\|(\tilde{\ell}_t-m_t)\|^2}$ variance term. However, with general second order corrections ($a_t \geq 0$), the key variance term is $\inn{u,a_t}$ as it corresponds to the best arm's second moment under a suitably chosen $u$ and the responsibility of cancelling the entire first term of \eqref{eqn:critical} now falls upon $\inn{x_t, a_t}$. Under limited information, negative entropy is unable to maintain this and we therefore have to incorporate the log barrier function (also see Lemma 1 in \citet{weiluo}). We now state our main result for adaptive label efficient prediction which bounds the regret of Algorithm \ref{algo:main2}.

\begin{theorem}\label{thm:main}
For $a_t = 6\eta \epsilon^2 (\tilde{\ell}_t-m_t)^2 $, $\tilde{\ell}_t = \frac{(\ell_t - m_t)}{\epsilon}\mathbbm{1}_{\{d_t=1\}} + m_t$, $\epsilon = \nicefrac{n}{T}$ and $\eta \leq \nicefrac{1}{162K}$ where the sequence of messages $m_t$ are generated using the reservoir sampling scheme, the expected regret of algorithm \ref{algo:main2} satisfies the following:
\[ \mathbb{E} \left[ R_T \right] \leq   \frac{\log K + \log T}{\epsilon\eta} + 18 \eta Q^* ~. \]
Furthermore, if $\epsilon Q^* \geq 1458K^2 \log KT $, then $ \mathbb{E} \left[ R_T \right] = \order{\left(\sqrt{\frac{Q^*T\log K}{n}}\right)}$ with an optimal choice of $\eta$.
\end{theorem}
\noindent
Consider a concrete example of a game played for time $T$, where we anticipate $Q^* \approx \sqrt{T}$ and $n \approx \sqrt{T}$. In this scenario, if we were to run the standard label efficient prediction algorithm as given in \citet{nicololep}, we would get a regret bound of $\order{\left(T^{\nicefrac{3}{4}}\right)}$; following an FTRL with negative entropy\footnote{As done in \citet{steinliang} for prediction with experts}-based strategy would be inapplicable in this setting due to the constraint we highlight in section \ref{sec:alg}, however, Algorithm \ref{algo:main2} would incur $\sqrt{T}$ regret -- a marked improvement. Also, note that because of the full vector feedback, it is not required to allocate any rounds \textit{exclusively} for reservoir sampling. This fact is reflected in not having to incur any additive penalty for reservoir sampling.  

Theorem \ref{thm:main} is slightly restricted in scope, due to the lower bound required on $\epsilon Q^*$, in its ability to attain the optimal regret scaling with quadratic variation. We now proceed to discuss what can be said without any constraint on $\epsilon Q^*$. Specifically, we will provide an algorithm obtaining $\order{(\sqrt{\nicefrac{QT\log K}{n}})}$ regret under \textit{all} scenarios, the trade-off however being that we will be penalized by $Q$ instead of $Q^*$. In settings where the $\epsilon Q^*$ condition does not hold and incurring regret in terms of $Q$ is not unfavourable (as an extreme example, consider constant variation on all arms, with very limited feedback) the strategy below will certainly be of use. The algorithm, again based on OMD, foregoes second order corrections and has updates defined by:
\begin{align}
x_t &= \argmin_{x \in \Omega} \{ \inn{x, \epsilon m_t}+D_\R(x,x_t^\p) \} \label{eqn:update3}\\ 
x_{t+1}^\p &= \argmin_{x \in \Omega} \{ \inn{x, \epsilon\tilde{\ell}_t}+D_\R(x,x_t^\p) \} \label{eqn:update4}
\end{align}
\noindent
Without second order corrections, the $\epsilon$ term can be folded into the regularizer and the updates reduce to the ones studied in \citet{rakhlinsridharan}. For updates \eqref{eqn:update3} and \eqref{eqn:update4}, we have the following analogue of Lemma \ref{thm:generalthm}, and then consequently, the analogue of Theorem \ref{thm:main}. We include these here in the interest of completeness, but equivalent statements can be found in \citet{rakhlinsridharan}.
\begin{lemma}\label{lemma:main2}
For any $u \in \Omega$, updates \eqref{eqn:update3} and \eqref{eqn:update4} guarantee that: 
\begin{align*}
&\inn{x_t-u, \tilde{\ell}_t} \leq \frac{1}{\epsilon} \bigg( D_\R(u,x_t^\p)-D_\R(u,x^\p_{t+1}) \\
&+ \inn{x_t - x'_{t+1}, \epsilon\tilde{\ell}_t - \epsilon m_t} - D_{\R}(x_{t+1}^\p, x_t)-D_{\R}(x_t, x_t^\p) \bigg)~.
\end{align*}

\end{lemma}

\begin{theorem}\label{thm:rakhsrimod}
For $\R = \frac{1}{\eta}\sum_{i=1}^K x_i \log x_i$, $\tilde{\ell}_t = \frac{(\ell_t - m_t)}{\epsilon}\mathbbm{1}_{\{d_t=1\}} + m_t$, $\epsilon = \nicefrac{n}{T}$ and $\eta > 0$, where the sequence of messages are generated using the reservoir sampling scheme, Algorithm \ref{algo:main2} with $a_t = 0$ yields:
\[ \Ex[R_T]  \leq \frac{\log K}{\eta \epsilon} +  \frac{\eta Q}{2}~. \]
Optimally tuning $\eta$ yields a $\order{\left( \sqrt{\nicefrac{QT\log K}{n}}\right)}$ bound.
\end{theorem}
\noindent
Trying to deeper understand how the constraint of Theorem \ref{thm:main} can be sidestepped to yield a universal algorithm dependent on $Q^*$ remains a direction of future interest.

\paragraph{Parameter-Free Algorithms}
Note that we have assumed knowledge of $T$, $Q$ and $Q^*$ when optimising for the fixed step size $\eta$ in the above discussion. This is often not possible and we now briefly discuss the extent to which we can obtain parameter-free algorithms. In Theorem \ref{thm:doubling} we claim that we can choose $\eta$ adaptively for the $Q$ dependent bound we present in Theorem \ref{thm:rakhsrimod} and discuss this in Appendix \ref{app:doubling}\footnote{Note that similarly to \citet{hazankale} we still assume knowledge of $T$, but this can be circumvented using standard tricks.}. It remains open whether a $Q^*$ dependent bound (or in general, any non-monotone dependent bound) can be made parameter free for even the standard prediction with expert advice problem. The challenge is essentially that our primary tool to sidestep prior knowledge of a parameter, the doubling trick, is inapplicable for non-monotone quantities.

Even freeing algorithms from prior knowledge of non-decreasing arm dependent quantities, such as $\max_i Q_i$ remains open for limited information setups (i.e. anything outside prediction with expert advice)  due to the lack of a clear auxiliary term one can observe. In Algorithm \ref{algo:doubling}, we proceed in epochs (or rounds) such that $\eta$ remains fixed per epoch. Denote by $\eta_\alpha$ the value of $\eta$ in epoch $\alpha$. We will write $T_\alpha$ for the first time instance in epoch $\alpha$.

\begin{theorem}\label{thm:doubling}
For the conditions mentioned in Theorem \ref{thm:rakhsrimod}, Algorithm \ref{algo:doubling} (a parameter free algorithm) achieves:

\[ \Ex[R_T] \leq \order{\left( \sqrt{\nicefrac{QT\log K}{n}}+ \sqrt{\log K} \right)}~. \]
\end{theorem}

\begin{algorithm}[H]
\caption{\textsc{Parameter Free Adaptive Label Efficient Prediction}}\label{algo:doubling}
\begin{algorithmic}[1]
\State \textbf{Initialize:} $\eta=\frac{\sqrt{2 \log K}}{\epsilon}, T_1=1, t=1.$
\For{$\alpha = 1, 2,\dots$}
	\State $x_{t}^\p=\argmin_{x\in \Omega}\R(x)$
	\While{$t \leq T$}
	\State Draw $d_t \sim \Ber(\epsilon)$, update $x_t$ according to \eqref{eqn:update3}
	\State Play $i_t\sim x_t$ and if $d_t=1$, observe $\ell_t$
	\State Update $x_{t+1}^\p$ according to \eqref{eqn:update4}
	\If{$\sum_{s=T_\alpha}^{t} \sum_{i=1}^K (\tilde{\ell}_{s,i}-m_{s,i})^2 \geq \frac{2\log K}{\epsilon^2 \eta_{\alpha -1}^2}$}
		\State $\eta \leftarrow \eta/2$, $T_{\alpha+1} \leftarrow t$, $t\leftarrow t+1$
		\State \textbf{break}
	\EndIf
	\State $t\leftarrow t+1$
	\EndWhile
\EndFor
\end{algorithmic}
\end{algorithm}

\section{Adapting to Slowly Varying Losses in Other Information-Constrained Games} \label{sec:extensions}

We will now investigate exploiting the regularity of losses in a variety of other settings with implicit/explicit information constraints. We will first  focus on bandit feedback, following which we will briefly discuss partial monitoring. The proofs for this section can be found in the supplementary material. 

\subsection{Label Efficient Bandits}\label{sec:leb}

The change here is in the feedback information the learner receives when asking for information. Instead of receiving the full loss vector, the learner now only receives the loss of the played arm $i_t$, i.e. the $i_t$th coordinate of $\ell_t$. We will continue to use the same update rules \eqref{eqn:update1} and \eqref{eqn:update2} here. What will change most importantly is the regularizer which will now solely be the log barrier regularizer $\R = \frac{1}{\eta} \sik \log \frac{1}{x_i}$. Note that the coefficient of log barrier is also $\nicefrac{1}{\eta}$ instead of the earlier $\nicefrac{1}{\eta K}$. The loss estimates and second order corrections will also change and these are all mentioned in Table \ref{table:directory}. We will now state the main theorem for label efficient bandits. Most of the analysis is similar to Theorem \ref{thm:main}, but we do highlight the differences in Appendix \ref{app:ext:leb} in the supplementary material. 
\begin{theorem}\label{thm:leb}
For $a_{t,i} = 6\eta \epsilon^2 x_{t,i} (\tilde{\ell}_t-m_t)^2 $, $\tilde{\ell}_t = \frac{\ell_t - m_t}{\epsilon x_{t,i}}\mathbbm{1}_{\{d_t=1, i_t = i\}} + m_{t,i}$, $\epsilon = \nicefrac{n}{T}$ and $\eta \leq \nicefrac{1}{162K}$ where the sequence of messages $m_t$ are given by reservoir sampling, the regret of algorithm \ref{algo:main2} modified for label efficient bandits satisfies: 
\[\mathbb{E} \left[ R_T \right] \leq   \frac{K\log T}{\epsilon\eta} + 18 \eta Q^* + K(\log T)^2~.  \]
\end{theorem}
\noindent
Note that since we are in the bandit feedback setting, we now reserve certain rounds solely for reservoir sampling. This is reflected in the additive $K(\log T)^2$ term in regret. There are now $(\log T)^2$ rounds allotted to each of the $K$ arms, hence the term. There will also be a few minor changes in the algorithm primarily corresponding to the appropriate execution of reservoir sampling for bandit feedback.


\subsection{Partial Monitoring}

We will now discuss adaptivity in partial monitoring games. A partial monitoring game $G = (L,H)$ is defined by a pair $L$ and $H$ of $K \times N$ matrices. Both matrices are visible to the learner and the adversary. At each time $t$, the learner selects a row (or arm, action) $i_t \in [K]$ and the opponent chooses a column $y_t \in [N]$. The learner then incurs a loss of $\ell(i_t, y_t)$ and observes feedback $h(i_t, y_t)$~\footnote{We are considering oblivious adversarial opponents as before and further take entries of $H$ to be in $[0,1]$. The assumption on the entries is not major since the learner can always appropriately encode the original entries by numbers.}. When clear from context, we will denote by $\ell(i,t)$ the loss of arm $i$ at time $t$ and by $h(i,t)$ the feedback of arm $i$ at time $t$. The expected regret here is: 
\[ \max_{i^* \in [K]} \mathbb{E} \left[ \sum_{t=1}^T \ell (i_t,y_t) - \sum_{t=1}^T \ell(i^*, y_t) \right]\]

\subsubsection{Revealing Action Partial Monitoring}\label{sec:revact}

First consider the class of partial monitoring games with a \textit{revealing action}-- that is, suppose $H$ has a row with $N$ distinct elements. It is clear that if the learner plays this row, they can receive full information regarding which column the adversary has chosen. The cost of playing this row very well defines which \textit{class} this game falls into (see for example the spam game discussed in \citet{nbd2}), but in general, the minimax regret of these games scales as $T^{\nicefrac{2}{3}}$ and these games therefore fall in the \textit{hard} class of games. Revealing action games and label efficient prediction differ in the way they \textit{charge} the learner for information. For label efficient prediction, we have seen that there is a fixed number of times (budget) one can obtain information, but there is no additional cost of doing so. In revealing action games however,  there is a loss associated to each time the learner asks for information. We will now show a reduction from this class of games to the standard label efficient prediction we discussed in sections \ref{sec:alg} and \ref{sec:analysis}.

Let the cost of playing the revealing action be $c = \max_{b \in [N]} L(a,b)$ where $a \in [K]$ is the revealing action row of $L$. Suppose $\alpha$ is the probability with which we play the revealing action at each round. $\alpha$ here corresponds to the $\epsilon$ from earlier sections, however $\alpha$ is now a free parameter\footnote{Note that the update rules \eqref{eqn:update1} and \eqref{eqn:update2} will now also have $\alpha$ in place of $\epsilon$}. We will still run reservoir sampling in the background as before to obtain the optimistic messages $m_t$. Now, in this light, the following theorem can be seen to follow from Theorem \ref{thm:main}.
\begin{theorem}
For $a_t = 6\eta \alpha^2 (\tilde{\ell}_t-m_t)^2 $, $\tilde{\ell}_t = \frac{(\ell_t - m_t)}{\alpha}\mathbbm{1}_{\{d_t=1\}} + m_t$, $\alpha \leq 1$ and $\eta \leq \nicefrac{1}{162K}$ where the sequence of messages $m_t$ are generated using reservoir sampling, the expected regret of algorithm \ref{algo:main2} modified for revealing action partial monitoring games with loss entries in $[0,1]$ satisfies the following:
\[ \mathbb{E} \left[ R_T \right] \leq   \frac{\log K + \log T}{\alpha\eta} + 18 \eta Q^* + \alpha Tc + (\log T)^2~. \] 
Optimising the parameters $\eta$ and $\gamma$ yields a bound of $\order{\left(\left(Q^*T \log K \right)^{\nicefrac{1}{3}}\right)}$~.
\end{theorem}
\noindent
Note that now, we will again have to allocate rounds specifically for reservoir sampling as was the case with bandits, hence the additive $(\log T)^2$ term. The added $\alpha Tc$ corresponds to the cost paid for playing the revealing action. 

\subsubsection{Hard Partial Monitoring Games}\label{sec:hardpm}

We now turn to the \textit{hard} class of partial monitoring games. As mentioned in \citet{piccolboni} and \citet{lugosibook}, we will assume that there exists a matrix $W$ such that $L = WH$. This is not an unreasonable assumption, as if this does not hold for the given $L$ and $H$, one can suitably modify (see \citet{piccolboni}) $L$ and $H$ to ensure $L' = W'H'$, and if this condition continues to fail after appropriate modifications, \citet{piccolboni} show that sublinear regret is not possible for the original $G = (L,H)$. Observe that $L = WH$ will allow us to write $\smash{\ell(i,t) = \sum_{j \in [K]} w(i,j)h(j,t)}$. Therefore:
\[ \tilde{\ell}(i,t) = \frac{\left(\sum_{j \in [K]}  w(i,j)h(j,t) - m_{t,i}\right)\mathbbm{1}_{\{i_t = j\}}}{x_{t,j}} + m_{t,i} \]
is now an unbiased estimate of  $\ell(i,t)$. $m_t$ is still the optimistic messages where $m_{t,i}$ corresponds to an estimate of the average loss incurred by arm $i$ till time $t$. These will still be obtained using reservoir sampling and we will maintain a separate reservoir for each arm $i \in [K]$. Note that since $\smash{\ell(i,t) = \sum_{j \in [K]} w(i,j)h(j,t)}$ and the matrices $L, W$, and $H$ are all visible to the learner, playing action $r$ at time $t$ for example will allow the learner to observe the $r$th component $w(i,r)h(r,t)$ of the loss for each action $i \in [K]$. Therefore, by maintaining an estimate (reservoir) for each \textit{component}, we will be able to maintain an estimate for each arm.

Now, for these games we will use optimistic OMD without second order corrections \citep{rakhlinsridharan, chiang}. The update rules are the same as equations \eqref{eqn:update3} and \eqref{eqn:update4} without the $\epsilon$ term. Additionally, the arm we play will be sampled from $w_t$ where $w_t = (1-\gamma)x_t + \gamma \one$. The forced exploration is necessary to allow a minimum mass on all arms. Note that the structure defined by $\ell(i,t) = \sum_{j \in [K]} w(i,j)h(j,t)$ says that we potentially have to play \textit{all} arms to maintain unbiased estimates of \textit{any} arm. This forced exploration is unavoidable (see \citet{lugosibook}).
\begin{theorem}\label{abc}
Given $G = (L,H)$ with loss entries in $[0,1]$, a matrix $W$ such that $L= WH$, $\eta >0$ and $\R = \nicefrac{1}{\eta}\sik x_i \log x_i$, the update rules \eqref{eqn:update3} and \eqref{eqn:update4} (omitting the $\epsilon$) mixed with $\gamma$ forced exploration satisfies:
$\Ex [R_T] \leq \frac{\log K}{\eta} + \frac{KQ\eta}{2\gamma} + \gamma T$. Optimising for $\eta$ and $\gamma$ gives us a regret of $\order{(\left(QT K \right)^{\nicefrac{1}{3}})}$~.
\end{theorem}
\noindent
Note here the strong dependence on $K$ which is an outcome of each $\ell (i,t)$ being dependent on potentially all ($K$) other actions.


\section{Lower Bounds}\label{sec:lowerbounds}

We now prove explicit quadratic variation-based lower bounds for (standard) label efficient prediction and label efficient bandits. By capturing both the constraint on information as well as the quadratic variation of the loss sequence, our lower bounds generalize and improve upon existing lower bounds. We extend the lower bounds for label efficient prediction to further incorporate the quadratic variation of the loss sequence and enhance the quadratic variation dependent lower bounds for multi-armed bandits to also include the constraint on information by bringing in the number of labels the learner can observe ($n$).

Our bounds will be proven in a 2-step manner similar to that in \citet{gerchinlatt}. The main feature of step 1 (the Lemma step) is that of centering the Bernoulli random variables around a parameter $\alpha$ instead of $\nicefrac{1}{2}$, which leads the regret bound to involve the $\alpha(1-\alpha)$ term corresponding to the variance of the Bernoulli distribution. Step 2 (the Theorem step) builds upon step 1 and shows the existence of a loss sequence belonging to an $\alpha$-variation ball (defined below) which also incurs regret of the same order. Recall the quadratic variation for a given loss sequence: $ \smash{Q = \sum_{t=1}^T \norm{\ell_t - \mu_T}_2^2 \leq \nicefrac{TK}{4}}$. Now, for $\alpha \in [0,\nicefrac{1}{4}]$ define an $\alpha$-variation ball as: $\smash{\cV_{\alpha} \triangleq \{ \{\ell_t\}_{t=1}^T \,: \; \nicefrac{Q}{TK} \leq \alpha \}}$. All of the proofs for this section have been postponed to Appendix \ref{app:lowerbound} in the supplementary material.

Theorems \ref{thmlb:lep} and \ref{thmlb:leb}, after incorporating $Q \leq \alpha TK$ give us lower bounds of $\smash{\Omega{(\sqrt{\nicefrac{QT\log (K-1)}{Kn}})}}$ and $\smash{\Omega{(\sqrt{\nicefrac{QT}{n}})}}$ respectively. Our corresponding upper bounds are $\smash{\order{(\sqrt{\nicefrac{QT\log K}{n}})}}$ and $\smash{\order{(\sqrt{\nicefrac{QTK}{n}})}}$~.\footnote{We upper bound all of our $Q^*$ dependent upper bounds by $Q$ so as to consistently compare with the lower bounds. Note that $Q^*$ and $Q$ are in general incomparable and all that be said is that $Q^* \leq Q$.} Comparing the two tells us that our strategies are optimal in their dependence on $Q$ and on the constraint in information indicated by $n$. There is however a gap of $\sqrt{K}$~. This gap was mentioned in \citet{gerchinlatt} for the specific case of the multi-armed bandit problem, and was closed recently in \citet{sparsity}. Barring the easy to see $\sqrt{\nicefrac{Q \log K}{K}}$ lower bound for prediction with expert advice (which is also what Theorem \ref{thmlb:lep} translates to for $n=T$), we are unaware of other fundamental $Q$ based lower bounds for prediction with expert advice. The upper bounds for prediction with expert advice however are of $\smash{\order{(\sqrt{Q \log K})}}$ (\citep{extractingcertaintyhk}, \citep{steinliang} etc.), and this again suggests the $\sqrt{K}$ gap. Closing this for prediction with expert advice, label efficient prediction and for label efficient bandits remains open, as does the question of finding $Q^*$ dependent lower bounds.

\paragraph{Label Efficient Prediction (Full Information)}

As mentioned previously, the main difference here from the standard label efficient prediction lower bound proof \citep{nicololep} is that of centering the Bernoulli random variables around a parameter $\alpha$ which is responsible for ultimately bringing out the quadratic variation of the sequence. Our main statements for label efficient prediction are as follows.

\begin{lemma}\label{lemlb:lep}

Let $\alpha \in (0,1)$, $K \geq 2$, $T \geq n \geq \frac{c^2 \log (K-1)}{1-\alpha} $. Then, for any randomized strategy for the label efficient prediction problem, there exists a loss sequence under which $ \mathbb{E} [R_T] \geq c T \sqrt{\frac{\alpha (1-\alpha)\log (K-1)}{n}} $ for ~$c = \nicefrac{\sqrt{e}}{\sqrt{5(1+e)}}$ where expectation is taken with respect to the internal randomization available to the algorithm and the random loss sequence. 
\end{lemma}

\begin{theorem}\label{thmlb:lep}

Let $K \geq 2$, $T \geq n \geq \max\{ 32\log (K-1),~ 256 \log T\}$ and $\alpha \in \left[ \max\left\{ \frac{32 \log T}{n}, \frac{8 \log (K-1)}{n} \right\}, \frac{1}{4} \right]$. Then, for any randomized strategy for the label efficient prediction problem, $\max_{\{\ell_{t}\} \in v_{\alpha}} \Ex[R_T] \geq 0.36 T\sqrt{\frac{\alpha  \log (K-1)}{n}}$ where expectation is taken with respect to the internal randomization available to the algorithm.
\end{theorem}


\paragraph{Label Efficient Bandits}

The main difference here from standard bandit proofs is that now, the total number of revealed labels (each label is now a single loss vector entry) cannot exceed $n$. Hence, the $\sum_{i \in [K]} N_i(t-1)$ term which appears in the analysis is upper bounded by $n$ (where  $N_i(t-1)$ denotes the pulls of arm $i$ up till time $t-1$).

\begin{lemma}\label{lemlb:leb}
Let $\alpha \in (0,1)$, $K \geq 2$, $T \geq n \geq K/(4(1-\alpha))$. Then, for any randomized strategy for the label efficient bandit problem, there exists a loss sequence under which
${\Ex[R_T] \geq \frac{T}{8}\sqrt{\alpha (1-\alpha) K/n}}$ where expectation is taken with respect to the internal randomization available to the algorithm and the random loss sequence. 
\end{lemma}

\begin{theorem} \label{thmlb:leb}

Let $K \geq 2$, $T \geq n \geq \max \{32K,~ 384 \log T  \} $ and $\alpha \in \left[ \max\left\{ \frac{2c \log T}{n}, \frac{8K}{n} \right\}, \frac{1}{4} \right]$ with $c = (4/9)^2 (3\sqrt{5}+1)^2 \leq 12$. Then, for any randomized strategy for the label efficient bandit problem, $\max_{\{\ell_{t}\} \in v_{\alpha}} \Ex[R_T] \geq 0.04 T\sqrt{\frac{\alpha  K}{n}}$ where expectation is taken with respect to the internal randomization available to the algorithm.

\end{theorem}


\section{Conclusion}

We consider problems lying at the intersection of 2 relevant questions in online learning---how does one adapt to slowly varying data, and what best can be done with a constraint on information. As far as we know, the proposed algorithms are the first to jointly address both of these questions. There remain plenty of open problems in the area. Seeing to what extent universal $Q^*$ dependent algorithms can be obtained in starved information settings is a direction of future interest, as is closing the gap in $K$ highlighted in Section \ref{sec:lowerbounds}. Moreover, extending the notion of adaptivity to partial monitoring games to consider locally observable games and even more interestingly, locally observable sub-games within hard games also remains open. Higher order lower bounds for partial monitoring games have also not been studied and one wonders to what extent adaptivity can help in partial monitoring.

\bibliography{lep_arxiv_v2}
\bibliographystyle{plainnat}

\newpage

\appendix


\vspace{100cm}
\section{Label Efficient Prediction Main Proofs}\label{app:main}

In this section, we will prove Lemma \ref{thm:generalthm} and Theorem \ref{thm:main}. \\

\begin{pf}[Proof of Lemma \ref{thm:generalthm}]
Let $\Omega$ be a convex compact set in $\mathbb{R}^K$, $\R$ be a convex function on $\Omega$, $x'$ be an arbitrary point in $\Omega$, $c$ be any point in $\mathbb{R}^K$, and $x^*=\argmin_{x\in \Omega}\{\inn{x,c}+D_{\R}(x,x')\}$. Then, for any $u \in \Omega$, we have (see for example \citep{rakhsrid1}) :
\[ \inn{x^*-u, c}\leq D_{\R}(u,x')-D_\R(u,x^*)-D_{\R}(x^*,x') \]
Applying this on our update rules \eqref{eqn:update1} and \eqref{eqn:update2} gives us:

\begin{align}
\inn{x_t-x_{t+1}^\p, \epsilon m_t} \leq D_\R(x_{t+1}^\p, x_t^\p)-D_\R (x_{t+1}^\p, x_t)-D_\R(x_t, x_t^\p). \label{lemma main ineq 1}
\end{align}
and
\begin{align}
\inn{x_{t+1}^\p-u, \epsilon \tilde{\ell}_t+ a_t} \leq D_\R(u,x_{t}^\p)-D_\R(u,x_{t+1}^\p)-D_\R(x_{t+1}^\p, x_{t}^\p);  \label{lemma main ineq 2}
\end{align}
where we chose $u = x'_{t+1}$ when applying it to update rule \eqref{eqn:update1}. Now observe that:
\begin{align}
&\inn{x_t-u, \epsilon \tilde{\ell}_t}\nonumber \\
&=\inn{x_t-u, \epsilon \tilde{\ell}_t+ a_t}-\inn{x_t, a_t}+\inn{u,  a_t}\nonumber \\
&=\inn{x_t-x_{t+1}^\p, \epsilon\tilde{\ell}_t+a_t}-\inn{x_t, a_t}+\inn{x_{t+1}^\p-u, \epsilon\tilde{\ell}_t+a_t}+\inn{u,a_t} \nonumber \\
&=\inn{x_t-x_{t+1}^\p, \epsilon\tilde{\ell}_t+a_t-\epsilon m_t}-\inn{x_t,a_t}+\inn{x_{t+1}^\p-u, \epsilon\tilde{\ell}_t+ a_t}+\inn{x_t-x_{t+1}^\p, \epsilon m_t}+\inn{u,a_t}  \label{lemmamaineq}
\end{align}
Combining the above inequalities with equation \eqref{eqn:critical} gives us
\begin{align}
\inn{x_t-u, \tilde{\ell}_t}\leq \frac{1}{\epsilon} \left( D_\R(u,x_t^\p)-D_\R(u,x^\p_{t+1})+\inn{u,a_t} - P_t \right),
\end{align}
where $P_t \triangleq D_{\R}(x_{t+1}^\p, x_t)+D_{\R}(x_t, x_t^\p)\geq 0$ (by non-negativity of Bregman divergence).
\end{pf}


\vspace{1cm}

\noindent
We will now proceed to prove a series of lemmas which will build towards the proof of Theorem \ref{thm:main} (see also Appendix B in \citet{weiluo})

\begin{lemma}\label{lem:stability1}
For some radius $r > 0$, define the ellipsoid $\mathcal{E}_{x}(r)=\left\{u \in \mathbb{R}^K : \norm{u-x}_{x}\leq r \right\}$.
If $x^\p \in \mathcal{E}_x(1)$, $\eta \leq \frac{1}{81K}$ , then, for all $i \in [K]$, we have  
$\frac{x'_i}{x_i} \leq \nicefrac{10}{9}$. Additionally, $\norm{u}_{x^\p} \geq \frac{9}{10}\norm{u}_{x}$ for all $u \in \mathbb{R}^K$.
\end{lemma}

\begin{pf}[Proof of Lemma \ref{lem:stability1}]
As $x' \in \mathcal{E}_x(1) $, we can say that $\sik \frac{1}{\eta} (x'_i - x_i)^2 \left(\frac{1}{x_i} + \frac{1}{ K x_i^2} \right) \leq 1$ which further implies $\sum_{i=1}^K \frac{1}{\eta K} \frac{(x'_i - x_i)^2}{x_i^2} \leq 1$. Hence, we have $\frac{|x'_i -x_i|}{x_i} \leq \sqrt{\eta K} \:~ \forall i$. Now, since $\eta \leq \frac{1}{81K}$, the first part of the lemma follows. Further observe $\norm{u}_{x^\p} = \sqrt{\frac{1}{\eta} \sik u_i^2 \left(\frac{1}{x'_i } + \frac{1}{ Kx'^2_i}\right)} \geq \frac{1}{\nicefrac{10}{9}}\norm{u}_{x} = \frac{9}{10}\norm{u}_{x}$~.
\end{pf}

\vspace{1cm}

\begin{lemma}\label{lem:conditionforstab}
Let $x_t$ and $x'_t$ correspond to our update rules \eqref{eqn:update1} and \eqref{eqn:update2} and suppose $\eta \leq \frac{1}{81K}$. Then, if $\norm{\epsilon(\tilde{\ell}_t-m_t)+a_t}_{x_t,*}\leq \frac{1}{3}$, we have $x^\p \in \mathcal{E}_x(1)$.
\end{lemma}

\begin{pf}[Proof of Lemma \ref{lem:conditionforstab}]
Let us rewrite our update rules \eqref{eqn:update1} and \eqref{eqn:update2} in the following way: 
\[ x_t = \argmin_{x \in \Omega} F_t(x) \text{ where } F_t(x) = \left\{ \inn{x, \epsilon m_t}+D_{\R}(x,x_t^\p) \right\} \]\vspace{-0.3cm}
\[ x_{t+1}^\p = \argmin_{x \in \Omega} F'_{t+1}(x) \text{ where } F'_{t+1}(x) = \left\{ \inn{x, \epsilon\tilde{\ell}_t+a_t}+D_{\R}(x,x_t^\p) \right\} \]
\noindent
Because of the convexity of $F'_t$~, to prove our claim, it is sufficient to show that $ F'_{t+1}(u) \geq F'_{t+1} (x_t)$ for all points $u$ on the boundary of the ellipsoid. By Taylor's theorem, we know that $\exists \: \xi $  on the line segment between $u$ and $x_t$ such that:

\begin{align*}
F^\p_{t+1}(u) &= F^\p_{t+1}(x_t)+\inn{\nabla F^{\p}_{t+1} (x_t), u - x_t}+ \frac{1}{2}(u - x_t)^\top\nabla^2 F^\p_{t+1}(\xi)(u - x_t)  \\
&= F^\p_{t+1}(x_t)+ \inn{\epsilon(\tilde{\ell}_t-m_t)+a_t,u-x_t} +\inn{\nabla F_t (x_t), u-x_t}\\
&~~~~~+ \frac{1}{2}(u - x_t)^\top\nabla^2 \R(\xi)(u - x_t) \\
&\geq F^\p_{t+1}(x_t)+ \inn{\epsilon(\tilde{\ell}_t-m_t)+a_t, u-x_t} + \frac{1}{2}\norm{u - x_t}_{\xi}^2 \\
&\geq F^\p_{t+1}(x_t)+ \inn{\epsilon(\tilde{\ell}_t-m_t)+a_t, u-x_t} + \frac{81}{200}\norm{u - x_t}_{x_t}^2  \\
&\geq F^\p_{t+1}(x_t)- \norm{\epsilon(\tilde{\ell}_t-m_t)+a_t}_{x_t,*} \norm{u - x_t}_{x_t} + \frac{1}{3}\norm{u - x_t}_{x_t}^2 \\
&= F^\p_{t+1}(x_t)- \norm{\epsilon(\tilde{\ell}_t-m_t)+a_t}_{x_t,*} + \frac{1}{3} \tag{as $\norm{u - x_t}_{x_t} =1$}\\
&\geq F^\p_{t+1}(x_t). 
\end{align*}
Where the first inequality follows from the optimality of $x_t$, the second from Lemma \eqref{lem:stability1}, the third from H\"older's inequality, and the last by the assumption of this lemma.

\end{pf}

\vspace{1cm}

\begin{lemma}\label{lem:correlation}
Let $x_t$ and $x'_t$ be defined as in our update rules \eqref{eqn:update1} and \eqref{eqn:update2}. Additionally, suppose $\eta \leq \frac{1}{81K}$. Then, if $\norm{\epsilon(\tilde{\ell}_t-m_t)+a_t}_{x_t,*} \leq \frac{1}{3}$, we have that $\norm{x_{t+1}^\p-x_t}_{x_t}\leq 3 \norm{\epsilon(\tilde{\ell}_t-m_t)+a_t}_{x_t,*}$~.
\end{lemma}

\begin{pf}[Proof of Lemma \ref{lem:correlation}]
We will begin by defining $F_t(x)$ and $F_{t+1}^\p(x)$ as above. Then we have that:

\begin{align}
F_{t+1}^\p(x_t)-F_{t+1}^\p(x_{t+1}^\p)&=\inn{x_t-x_{t+1}^\p,~ \epsilon(\tilde{\ell}_t-m_t) +a_t} + F_t(x_t)-F_t(x_{t+1}^\p) \nonumber \\
&\leq \inn{x_t-x_{t+1}^\p,~\epsilon(\tilde{\ell}_t-m_t)+a_t} \nonumber \\
&\leq \norm{x_t-x_{t+1}^\p}_{x_t}\norm{\epsilon(\tilde{\ell}_t-m_t)+a_t}_{x_t,*} \label{eq:ab1}
\end{align}
\noindent
By Taylor's theorem and the optimality of $x_{t+1}^\p$, we again have that,

\begin{align}
F_{t+1}^\p(x_t)-F_{t+1}^\p(x_{t+1}^\p)&=\inn{\nabla F_{t+1}^\p(x_{t+1}^\p), x_t-x_{t+1}^\p} + \frac{1}{2}(x_t-x_{t+1}^\p)^\top \nabla^2 F_{t+1}^\p(\xi)(x_t-x_{t+1}^\p) \nonumber \\
&\geq \frac{1}{2}\norm{x_t-x_{t+1}^\p}_{\xi}^2  \nonumber \\
&\geq \frac{1}{3}\norm{x_t-x_{t+1}^\p}_{x_t}^2 \label{eq:ab2}
\end{align}
\noindent
where the last inequality again follows using the same arguments as done in Lemma \eqref{lem:conditionforstab}. Combining \eqref{eq:ab1} and \eqref{eq:ab2} proves the claimed result.

\end{pf}

\vspace{1cm}

\begin{lemma}\label{lem:almost}
For $a_t = 6\eta\epsilon^2(\tilde{\ell}_{t}-m_{t})^2$, \: $\tilde{\ell}_{t} = \frac{(\ell_t - m_t)}{\epsilon}\mathbbm{1}_{\{ d_t =1 \}} + m_t$,~ $\eta \leq \frac{1}{162K}$  we have that $\norm{\epsilon(\tilde{\ell}_t-m_t)+a_t}_{x_t,*}\leq \frac{1}{3}$.
\end{lemma}

\begin{pf}[Proof of Lemma \ref{lem:almost}]

\begin{align*}
\norm{\epsilon(\tilde{\ell}_t-m_t)+a_t}^2_{x_t,*}
&= \sik \frac{\eta}{ \frac{1}{x_i} + \frac{1}{ K x_i^2}} \left(\epsilon(\tilde{\ell}_{t,i}-m_{t,i}) + 6 \eta \epsilon^2 (\tilde{\ell}_{t,i}- m_{t,i})^2 \right)^2\\
&= \eta \sik \frac{\epsilon^2(\tilde{\ell}_{t,i}- m_{t,i})^2}{\frac{1}{x_i} + \frac{1}{ K x_i^2}} \left[ 1 + 12\eta\epsilon (\tilde{\ell}_{t,i}- m_{t,i}) + 36\eta ^2 \epsilon^2 (\tilde{\ell}_{t,i}- m_{t,i})^2\right]\\
&\leq 2\eta \epsilon^2 \sik (\tilde{\ell}_{t,i}- m_{t,i})^2 x_i\\
&\leq 2 \eta\\
&\leq \frac{1}{9} 
\end{align*}
\noindent
The above inequalities follow by observing that $|\epsilon(\tilde{\ell}_{t,i}-m_{t,i})| = |(\ell_t - m_t)\mathbbm{1}_{\{ d_t =1 \}}| \leq 1$ along with the assumption on $\eta$.

\end{pf}

\vspace{1cm}
\noindent
We are now in a position to prove Theorem \ref{thm:main}.

\begin{pf}[Proof of Theorem \ref{thm:main}]
We will first show that our choices of loss vectors, messages, and corrections obey the condition of Lemma \ref{thm:generalthm}. To this end, observe that:

\begin{align*}
\inn{x_t-x_{t+1}^\p, \epsilon(\tilde{\ell}_t-m_t)+ a_t}
&\leq \norm{x_t-x_{t+1}^\p}_{x_t}\norm{\epsilon(\tilde{\ell}_t-m_t)+a_t}_{x_t,*}\\
&\leq 3\norm{\epsilon(\tilde{\ell}_t-m_t)+a_t}_{x_t,*}^{2}\\
&\leq 3\eta \sik \frac{\epsilon^2(\tilde{\ell}_t- m_t)^2}{\frac{1}{x_i} + \frac{1}{ K x_i^2}} \left[ 1 + 12\eta\epsilon (\tilde{\ell}_t- m_t) + 36 \eta ^2 \epsilon^2 (\tilde{\ell}_t- m_t)^2\right] \\
&\leq 6 \eta \epsilon^2 \sik x_{t,i}(\tilde{\ell}_{t,i}-m_{t,i})^2 = \inn{x_t, a_t}
\end{align*}
where the first inequality follows from H\"older's inequality, the second from Lemma \eqref{lem:correlation}, and the last 2 are as done in the proof of Lemma \ref{lem:almost}. We can therefore proceed to sum both sides of the result of Lemma \ref{thm:generalthm} over $t$ to get: 
\begin{align*}
\mathbb{E} \left[ \sum_{t=1}^T \inn{x_t-u, \tilde{\ell}_t} \right] &\leq \frac{1}{\epsilon} \sum_{t=1}^T \mathbb{E}\left[ \left( D_\R(u,x_t^\p)-D_\R(u,x^\p_{t+1})+\inn{u,a_t} \right) \right]
\end{align*}
Now we can see that the first 2 terms on the right hand side will telescope to yield a remaining term of $D_\R (u, x'_1)$. We will pick $u = (1 - \frac{1}{T})\basis_{i^*} + \frac{1}{KT}\one$ instead of simply $\basis_{i^*}$ so as to ensure that the log barrier component is well defined. Hence we will have:
\begin{align}
\mathbb{E} \left[ \sum_{t=1}^T \inn{x_t-u, \tilde{\ell}_t}\right] &\leq \frac{1}{\epsilon} \left( D_\R (u, x'_1) + \mathbb{E} \left[ \sum_{t=1}^T \inn{u,a_t}   \right] \right) \label{aaaa}
\end{align}
\begin{align*}
D_\R(u, x_1') &= \R(u) - \R(x_1') - \inn{\nabla \R(x_1'), u - x_1'} \\
&= \R(u) - \R(x_1') \leq \frac{\log T}{\eta} + \frac{\log K}{\eta} = \frac{\log K + \log T}{\eta}
\end{align*}
This choice of $u$ will also introduce an additional term in regret of $\Ex  \frac{1}{T} \sum_{t=1}^T \inn{ x'_1 - e_{i^*} , \tilde{\ell}_t + a_t }$, but as can be seen in \citet{weiluo}, this term is $\order{(1)}$~.
\begin{align}
&\mathbb{E} \left[ \sum_{t=1}^T \inn{u,a_t}   \right] = 6\eta \epsilon^2 \mathbb{E} \left[ \sum_{t=1}^T \frac{(\ell_{t,i^*} - m_{t,i^*})^2}{\epsilon^2}\mathbbm{1}_{\{ d_t =1 \}} \right]\\
&\leq 18\eta \epsilon \left[ \sum_{t=1}^T(\ell_{t,i^*}-\mu_{t,i^*})^2+\sum_{t=1}^T(\mu_{t,i^*}-\mu_{t-1,i^*})^2 + \mathbb{E}\left[\sum_{t=1}^T(\mu_{t-1,i^*}-\tilde{\mu}_{t-1,i^*})^2\right]\right] \label{eqn:aa}
\end{align}
The first and third terms of \eqref{eqn:aa} can be bounded using Lemmas 10 and 11 from \citep{hazankale} and are order $\order{(Q_{T,i^*}+1)}$. The middle term above is $\order{(1)}$ from Lemma 18 in \citep{weiluo}. Therefore, substituting everything back into \eqref{aaaa}, we have that:
\begin{align}
\mathbb{E} \left[ R_T \right] \leq \frac{\log K + \log T}{\epsilon\eta} + 18 \eta Q^* 
\end{align}
\end{pf}

\begin{pf}[Proof of Lemma \ref{lemma:main2}]
We will proceed similarly to the proof of Lemma \ref{thm:main} and rewrite \eqref{lemmamaineq} with $a_t =0$ :

\[\inn{x_t-u, \epsilon \tilde{\ell}_t} = \inn{x_t-x_{t+1}^\p, \epsilon\tilde{\ell}_t -\epsilon m_t}+\inn{x_{t+1}^\p-u, \epsilon\tilde{\ell}_t}+\inn{x_t-x_{t+1}^\p, \epsilon m_t} \]
We will again use the inequalities \eqref{lemma main ineq 1} and \eqref{lemma main ineq 2} (with $a_t =0$) to get:
\begin{align*}
\inn{x_t-u, \epsilon \tilde{\ell}_t} &\leq \inn{x_t-x_{t+1}^\p, \epsilon\tilde{\ell}_t -\epsilon m_t}+ D_\R(u,x_{t}^\p)-D_\R(u,x_{t+1}^\p)-D_\R (x_{t+1}^\p, x_t)-D_\R(x_t, x_t^\p)
\end{align*}
which proves the lemma after rearranging the $\epsilon$~.

\end{pf}

\begin{pf}[Proof of Theorem \ref{thm:rakhsrimod}]
\begin{align*}
\epsilon\inn{x_t-u, \tilde{\ell}_t} &\leq   D_\R(u,x_t^\p)-D_\R(u,x^\p_{t+1})+ \inn{x_t - x'_{t+1}, \epsilon\tilde{\ell}_t - \epsilon m_t} - D_{\R}(x_{t+1}^\p, x_t)-D_{\R}(x_t, x_t^\p) \\
&\leq D_\R(u,x_t^\p)-D_\R(u,x^\p_{t+1}) + \frac{1}{2\eta}\|x_t - x'_{t+1}\|_2^2 + \frac{\eta}{2} \|\epsilon\tilde{\ell}_t - \epsilon m_t\|_2^2 - D_{\R}(x_{t+1}^\p, x_t)\\
&\leq  D_\R(u,x_t^\p)-D_\R(u,x^\p_{t+1}) + \frac{\eta \epsilon^2}{2}\|\tilde{\ell}_t -  m_t\|_2^2
\end{align*}
where the first inequality follows from Lemma \ref{lemma:main2}, the second one follows from H\"older's inequality and the non-negativity of the Bregman divergence, and the final one from the strong convexity of negative entropy in the $\ell_2$ norm.
We therefore have that $\inn{x_t-u, \tilde{\ell}_t} \leq \frac{1}{\epsilon} \big( D_\R(u,x_t^\p)-D_\R(u,x^\p_{t+1}) + \frac{\eta \epsilon^2}{2}\|\tilde{\ell}_t -  m_t\|_2^2 \big)$~. Now summing both sides over $t$ will yield:
\begin{align}
 \sum_{t=1}^T \inn{x_t-u, \tilde{\ell}_t} &\leq \frac{D_\R(u,x_1^\p)}{\epsilon} + \frac{\eta\epsilon}{2} \sum_{t=1}^T  \|\tilde{\ell}_t -  m_t\|_2^2\\
& \leq \frac{\log K}{\eta \epsilon} + \frac{\eta\epsilon }{2}\sum_{t=1}^T \sum_{i=1}^K  (\tilde{\ell}_{t,i} - m_{t,i})^2\label{eq:used in doubling}
\end{align}
Now, substituting the stated estimators, unravelling the right hand side similar to the analysis of \eqref{eqn:aa} and taking expectation will yield the $\frac{\log K}{\eta \epsilon} + \frac{\eta Q}{2}$ upper bound.

\end{pf}

\section{Parameter-Free Algorithms} \label{app:doubling}

We will proceed similarly to \citet{weiluo} in epochs where $\eta$ remains fixed per epoch. Let $\eta_\alpha$ be the value of $\eta$ in epoch $\alpha$. We write $T_\alpha$ for the first time instance in epoch $\alpha$. To simplify things, we still assume knowledge of $T$, however this assumption can be sidestepped.

\begin{pf}[Proof of Theorem \ref{thm:doubling}]
We start from \eqref{eq:used in doubling} and get the following for some epoch $\alpha$:
\begin{align*}
\sum_{t = T_{\alpha}}^{T_{{\alpha}+1}-1}\inn{x_t-u, \tilde{\ell}_t} &\leq \frac{1}{\epsilon} \left[ \frac{\log K}{\eta_\alpha} + \frac{\eta_\alpha\epsilon^2}{2}\sum_{t=1}^T \sum_{i=1}^K  (\tilde{\ell}_{t,i} - m_{t,i})^2 \right]\\
&= \order{\left( \frac{\log K}{\epsilon\eta_\alpha}\right)}
\end{align*}

\noindent
We can consequently write:
\[ \sum_{t=1}^T \inn{x_t-u, \tilde{\ell}_t} \leq \sum_{\alpha =0}^{\alpha^*} \order{\left(\frac{\log K }{\epsilon \eta_\alpha}\right)} \leq \order{\left( 2^{\alpha^*} \sqrt{\log K}\right)} \]
\noindent
where $\alpha^*$ is the epoch at $T$. Now we also know that epoch $\alpha^* -1$ has completed, hence:

\[  \sum_{t = T_{\alpha^*-1}}^{T_{\alpha^*}-1}\sum_{i =1}^K (\tilde{\ell}_{t,i}-m_{t,i})^2 \geq \frac{2\log K}{\epsilon^2 \eta_{\alpha^* -1}^2} = \Omega \left( 2^{2\alpha^*}\right)
\]
\noindent
So, we can write the entire bound as

\begin{align*}
\sum_{t=1}^T \inn{x_t-u, \tilde{\ell}_t} \leq \order{\left( 2^{\alpha^*} \sqrt{\log K}\right)} &\leq \order{\left( \sqrt{ \log K \sum_{t = T_{\alpha^*-1}}^{T_{\alpha^*}}\sum_{i =1}^K  (\tilde{\ell}_{t,i}-m_{t,i})^2} \right)}\\
&\leq \order{\left( \sqrt{\log K \sum_{t = 1}^{T}\sum_{i =1}^K  (\tilde{\ell}_{t,i}-m_{t,i})^2} \right)}
\end{align*}
\noindent
Also consider the case when $\alpha^* =0$, where $\sum_{t=1}^T \inn{x_t-u, \tilde{\ell}_t} \leq \sqrt{\log K}$~. Combining the above 2 cases, we get:

\[ \sum_{t=1}^T \inn{x_t-u, \tilde{\ell}_t} \leq \order{\left( \sqrt{\log K \sum_{t = 1}^{T}\sum_{i =1}^K  (\tilde{\ell}_{t,i}-m_{t,i})^2}+ \sqrt{\log K} \right)}  \]

\noindent
Taking expectation and using Jensen's inequality gives us:

\[ \Ex[R_T] \leq \order{\left( \sqrt{\log K~ \Ex\sum_{t = 1}^{T}\sum_{i =1}^K  (\tilde{\ell}_{t,i}-m_{t,i})^2}+ \sqrt{\log K} \right)} \]
\noindent
We can now plug in the usual $\tilde{\ell}_{t,i} =
\frac{(\ell_{t,i} - m_{t,i})}{\epsilon}\mathbbm{1}_{\{d_t=1\}} + m_{t,i}$, and choose messages corresponding to the quadratic variation based bound (i.e. $m_t = \tilde{\mu}_t$ via reservoir sampling) to give us:

\[ \Ex[R_T] \leq \order{\left( \sqrt{\nicefrac{QT\log K}{n}}+ \sqrt{\log K} \right)} \]
\noindent
Note that once again, taking expectation for the above estimates and messages will have to be done carefully similarly to as it is done for \eqref{eqn:aa}.

\end{pf}

\section{Proofs for Section \ref{sec:extensions}}\label{app:ext}

\subsection{Label Efficient Bandits} \label{app:ext:leb}

The sequence of lemmas for proving Theorem \ref{thmlb:leb} will be very similar as that done above for Theorem \ref{thm:main}. As mentioned previously, the key difference in the label efficient bandit setting is that we will have just the log barrier regularizer (instead of the hybrid regularizer). Additionally, our second order corrections are also $\smash{a_t = 6\eta\epsilon^2x_{t}(\tilde{\ell}_{t}-m_{t})^2}$. Lemmas \ref{lem:stability1}, \ref{lem:conditionforstab}, and \ref{lem:correlation} will follow almost identically. We provide the analogue to Lemma \ref{lem:almost} below and then prove Theorem \ref{thm:leb}.

\begin{lemma}\label{lem:leb1}
For $a_t = 6\eta\epsilon^2x_{t,i}(\tilde{\ell}_{t,i}-m_{t,i})^2$, \: $\tilde{\ell}_{t,i} = \frac{(\ell_{t,i}-m_{t,i})}{\epsilon x_{t,i}}\mathbbm{1}_{\{d_t =1, i_t=i\}} + m_{t,i}$, $\eta \leq \frac{1}{162K}$  we have that $\norm{\epsilon(\tilde{\ell}_t-m_t)+a_t}_{x_t,*}\leq \frac{1}{3}$.
\end{lemma}

\begin{pf}[Proof of Lemma \ref{lem:leb1}]

\begin{align*}
\norm{\epsilon(\tilde{\ell}_t-m_t)+a_t}^2_{x_t,*}
&= \sik \eta x_i^2 \left(\epsilon(\tilde{\ell}_{t,i}-m_{t,i}) + 6 \eta \epsilon^2 x_i (\tilde{\ell}_{t,i}- m_{t,i})^2 \right)^2\\
&= \eta \sik x_i^2 \epsilon^2 (\tilde{\ell}_{t,i}-m_{t,i})^2 \times\\
&~~~~~ \left[ 1 + 12\eta\epsilon x_i (\tilde{\ell}_{t,i}- m_{t,i}) + 36 \eta ^2 \epsilon^2 x_i^2 (\tilde{\ell}_{t,i}- m_{t,i})^2\right]\\
&\leq 2\eta \epsilon^2 \sik (\tilde{\ell}_{t,i}- m_{t,i})^2 x_i^2\\
&\leq 2 \eta\\
&\leq \frac{1}{9} 
\end{align*}
\noindent
The above inequalities again follow by observing that $|\epsilon x_{t,i}(\tilde{\ell}_{t,i}-m_{t,i})| = |(\ell_t - m_t)\mathbbm{1}_{\{ d_t =1, i_t =1 \}}| \leq 1$ along with the assumption on $\eta$.

\end{pf}


\begin{pf}[Proof of Theorem \ref{thm:leb}]
As before, we will again show that our choices of loss estimates, messages, and corrections guarantee Lemma \ref{thm:generalthm}. 

\begin{align*}
\inn{x_t-x_{t+1}^\p, \epsilon(\tilde{\ell}_t-m_t)+ a_t}
&\leq \norm{x_t-x_{t+1}^\p}_{x_t}\norm{\epsilon(\tilde{\ell}_t-m_t)+a_t}_{x_t,*}\\
&\leq 3\norm{\epsilon(\tilde{\ell}_t-m_t)+a_t}_{x_t,*}^{2}\\
&\leq 3\eta \sik \epsilon^2(\tilde{\ell}_{t,i}- m_{t,i})^2 x_i^2 \times \\
&~~~~~ \left[ 1 + 12\eta\epsilon x_i (\tilde{\ell}_{t,i}- m_{t,i}) + 36 \eta ^2 \epsilon^2 x_i^2 (\tilde{\ell}_{t,i}- m_{t,i})^2\right] \\
&\leq 6 \eta \epsilon^2 \sik x_{t,i}^2(\tilde{\ell}_{t,i}-m_{t,i})^2 = \inn{x_t, a_t}
\end{align*}
\noindent
Therefore we can again proceed to take summation over $t$ on both sides of the result of Lemma \ref{thm:generalthm}.

\begin{align*}
\mathbb{E} \left[ \sum_{t=1}^T \inn{x_t-u, \tilde{\ell}_t} \right] &\leq \frac{1}{\epsilon} \sum_{t=1}^T \mathbb{E}\left[ \left( D_\R(u,x_t^\p)-D_\R(u,x^\p_{t+1})+\inn{u,a_t} \right) \right]
\end{align*}
The first 2 terms on the right hand side will again telescope to yield a remaining $D_\R (u, x'_1)$, therefore giving us:
\begin{align*}
\mathbb{E} \left[ \sum_{t=1}^T \inn{x_t-u, \tilde{\ell}_t}\right] &\leq \frac{1}{\epsilon} \left( D_\R (u, x'_1) + \mathbb{E} \left[ \sum_{t=1}^T \inn{u,a_t}   \right] \right)
\end{align*}
\begin{align*}
D_\R(u, x_1') &= \R(u) - \R(x_1') - \inn{\nabla \R(x_1'), u - x_1'} \\
&= \R(u) - \R(x_1') \leq \frac{K\log T}{\eta} 
\end{align*}
Note that this time, we will not have the cancellation of $K$ as we did for Theorem \ref{thm:main}. We will pick $u = (1 - \frac{1}{T})\basis_{i^*} + \frac{1}{KT}\one$ as before. 
The rest of the proof will follow similarly to Theorem \ref{thm:main} to ultimately give us:
\begin{align}
\mathbb{E} \left[ \sum_{t=1}^T \inn{x_t-u, \tilde{\ell}_t}\right] = \order \left( \frac{K\log T}{\epsilon\eta} + 18 \eta Q^* + K(\log T)^2 \right)
\end{align}
Also note that now, we will have an added reservoir sampling cost in the final regret bound which is the $K(\log T)^2$ term.
\end{pf}

\subsection{Partial Monitoring}

\begin{pf}[Proof of Theorem \ref{abc}]
Following Lemma 2 from \citep{rakhlinsridharan} with $\R$ being standard negative entropy, gives us:
\begin{align*}
\Ex [R_T] &\leq \frac{\log K}{\eta} + \frac{\eta}{2} \sum_{t=1}^T \sum_{i=1}^K \Ex \left[ \frac{(\sum_{j \in [K]}  w(i,j)h(j,t) - m_{t,i})^2\mathbbm{1}_{\{i_t = j\}}}{x_{t,j}^2} \right] + \gamma T\\
&\leq \frac{\log K}{\eta} + \frac{K\eta}{2\gamma} \sum_{t=1}^T \sum_{i=1}^K \bigg(\sum_{j=1}^K  w(i,j)h(j,t) - m_{t,i}\bigg)^2 + \gamma T \leq \frac{\log K}{\eta} + \frac{KQ\eta}{2\gamma} + \gamma T
\end{align*}
We can therefore see that tuning $\eta$ and $\gamma$ gives us a bound of $\order{(\left(QT K \right)^{\nicefrac{1}{3}})}$. The second inequality follows by cancelling one $x_{t,j}$ with the indicator in expectation and bounding the remaining $x_{t,j}$ by the forced exploration. The last inequality follows from the definition of $\ell(i,t)$ and quadratic variation $Q$.
\end{pf}



\section{Lower Bound Proofs} \label{app:lowerbound}

\begin{pf}[Proof of Lemma \ref{lemlb:lep}]
Our proof for this lemma closely follows the proof of \citet{nicololep} with a few changes:
\begin{itemize}
\item Our Bernoulli random variables are centred at $\alpha$ instead of at $\nicefrac{1}{2}$.
\item We define our random variables a little differently to make the calculations easier. Namely, $Z^*$ is a Bernoull$(\alpha)$ random variable instead of Bernoulli $(\alpha - \epsilon)$ as is done in \citet{nicololep} (for $\alpha  = \nicefrac{1}{2}$) and $Z_j$ is Bernoulli $(\alpha + \epsilon)$ instead of Bernoulli $(\alpha)$ (again for $\alpha  = \nicefrac{1}{2}$).  
\end{itemize}
Given $y^t \in [0,1]$, consider the first $K$ coefficients of its unique dyadic expansion and denote these as $y^t_1, y^t_2,\dots, y^t_K$. We will then define $\ell_{t,i} = y^t_i$ for all $i \in [K] = \{ 1,2,\dots, K \}$. We will construct a random outcome sequence $Y_1,\dots, Y_T$, where each random variable is supported on $[0,1]$. The realizations of these random variables will then define an associated loss sequence as explained above. We will show that the expected regret of any randomized algorithm is bounded below by the claimed quantity, where we will take expectation with respect to the random outcome sequence as well as the internal/auxiliary randomness available to the algorithm. Denote by $A_1, A_2,\dots A_T$ the internal randomization available to the strategy (associated distribution is $\mathbb{P}_A$), which we will take to be an i.i.d. sequence of uniform random variables supported on $[0,1]$. Now define $K$(no. of arms) joint distributions $\mathbb{P}_i \otimes \mathbb{P}_A$ where $\PP_1,\dots, \PP_K$ are probability distributions over the outcome sequence which we define below. For $i \in [K]$, define by $\Q_i$ the distribution of: 

\[ Z^*2^{-i} + \sum_{j = 1,\dots, K,\ j \neq i} Z_j2^{-j} + 2^{-(K+1)}A \]
\noindent
$A, Z^*, Z_1,\dots Z_K$ are all independent random variables. $A$ is distributed uniformly over $[0,1]$, $Z^*$ is a Bernoulli $(\alpha)$ random variable, and $Z_j$ is distributed Bernoulli $(\alpha + \E)$ (we specify $\epsilon$ later).
\noindent
Now, under $\PP_i$~, the outcome sequence $Y_1,\dots, Y_T$ is i.i.d. from $ \Q_i$~. Hence, under $\PP_i$, for all $j \in [K]$ and $t \in [T]$, $\ell_{t,j}$ are i.i.d. Bernoulli random variables. $\ell_{t,i}$ is Bernoulli $(\alpha)$, and $\ell_{t,j}$, for $j \neq i$ is Bernoulli $(\alpha + \E)$.
\noindent
Denote the cumulative loss of the strategy by $\hat{L}_T = \sum_{t=1}^T \ell_{t,i_t}$ and the cumulative loss of arm $i$ by $L_{T,i}$. Let $\Ex_i$ be the expectation with respect to $\PP_i$ and $\Ex_A$ the expectation with respect to $\PP_A$. 
We then have that:
\begin{align*}
\max_{\{\ell_s\}_{s=1}^T} \left( \mathbb{E}_A \hat{L}_T - \min_{i \in [K]} L_{T,i} \right) &= \max_{\{\ell_s\}_{s=1}^T,~i \in [K]}  \left( \mathbb{E}_A \hat{L}_T - L_{T,i} \right)\\
&\geq \max_{i \in [K]} \mathbb{E}_i \left[ \mathbb{E}_A \hat{L}_T - L_{T,i} \right]
\end{align*}
Using Lemma \ref{lem:intermediate}, we have that $\PP_A \left[i_t =i|\{\ell_s\}_{s=1}^{t-1}\right] = \sum_{d=1}^D \beta_d \mathbbm{1}_{\left[i_t^d =i| \{\ell_s\}_{s=1}^{t-1}\right]}$ where $\mathbbm{1}_{\left[i_t^d =i| \{\ell_s\}_{s=1}^{t-1}\right]}$ is an indicator for the $d$-th deterministic algorithm choosing $i$. We therefore rewrite the regret as:
\begin{align*}
\max_{i \in [K]}\mathbb{E}_i \left[ \mathbb{E}_A \hat{L}_T - L_{T,i} \right] &= \max_{i \in [K]} \mathbb{E}_i \left[ \sum_{t=1}^T\sum_{d=1}^D \beta_d \sum_{k=1}^K \mathbbm{1}_{\left[i_t^d =i| \{\ell_s\}_{s=1}^{t-1}\right]} \ell_{t,k} - L_{T,i} \right]\\
&= \max_{i \in [K]} \sum_{d=1}^D \beta_d \mathbb{E}_i \left[ \sum_{t=1}^T \sum_{k=1}^K \mathbbm{1}_{\left[i_t^d =i| \{\ell_s\}_{s=1}^{t-1}\right]} \ell_{t,k} - L_{T,i} \right]\\
&= \mathcal{E} \max_{i \in [K]} \sum_{d=1}^D \beta_d \sum_{t=1}^T \PP_i \left[ i_t^d \neq i \right]\\
&= \mathcal{E}T \left( 1 - \min_{i \in [K]} \sum_{d=1}^D \sum_{t=1}^T \frac{\beta_d}{T} \PP_i \left[ i_t^d = i \right] \right)
\end{align*} 
where the third equality uses the fact that the regret grows by $\mathcal{E}$ under $\PP_i$ whenever $i_t \neq i$. Now for the $d$-th deterministic algorithm, let $1\leq T_1^d \leq \dots \leq T_n^d \leq T$ be the times when the strategy asks for the $n$ labels. Then $T_1^d,\dots, T_n^d$ correspond to the finite stopping times with respect to the i.i.d. process $Y_1,\dots Y_T$. Hence, the revealed outcomes $Y_{T_1^d},\dots, Y_{T_n^d}$ are i.i.d. from $Y_1$ (see \citet{chowteicher}). Denote by $R_t^d$ the number of revealed labels at time $t$. Now, as the subalgorithms are deterministic, $R_t^d$ is fully determined by $Y_{T_1^d},\dots, Y_{T_n^d}$. Hence, in general, $i_t^d$ can be thought to be a function of $Y_{T_1^d},\dots, Y_{T_n^d}$ instead of the revealed labels \textit{just} till time $t$, which are $Y_{T_1^d},\dots, Y_{T_{R_t^d}^d}$. As the joint distribution of $Y_{T_1^d},\dots, Y_{T_n^d}$ under $\PP_i$ is $\mathbb{Q}_i^n$, we have that $\PP_i[i_t^d =i] = \mathbb{Q}_i^n[i_t^d =i]$. Hence the regret becomes:
\begin{align*}
\max_{i \in [K]}\mathbb{E}_i \left[ \mathbb{E}_A \hat{L}_T - L_{T,i} \right] &= \mathcal{E}T \left( 1 - \min_{i \in [K]} \sum_{d=1}^D \sum_{t=1}^T \frac{\beta_d}{T} \mathbb{Q}_i \left[ i_t^d = i \right] \right)
\end{align*}
By the generalized Fano's inequality, we know that $\min_{i \in [K]} \sum_{d=1}^D \sum_{t=1}^T \frac{\beta_d}{T} \mathbb{Q}_i \left[ i_t^d = i \right] \leq \max \left\{ \frac{e}{1+e}, \frac{\bar{S}}{\log (K-1)} \right\}$ where $\bar{S} = \frac{1}{K-1} \sum_{i=2}^K \text{KL}(\Q_i^n, \Q_1^n)$. \\
Now observe that:
\begin{align*}
\text{KL}(\Q_i^n, \Q_1^n) &= n \text{KL}(\Q_i, \Q_1)\\
&\leq n(\text{KL}(\Ber(\alpha), \Ber(\alpha + \mathcal{E})) + \text{KL}(\Ber{(\alpha + \mathcal{E})}, \Ber(\alpha)))\\
&\leq n\left(\chi^2 (\alpha, \alpha + \E) + \chi^2 (\alpha + \E, \alpha)\right)\\
&= n\left( \frac{\E^2}{(\alpha + \E)(1-\alpha-\E)} + \frac{\E^2}{\alpha(1-\alpha)}\right)\\
&\leq \frac{5n\E^2}{\alpha(1-\alpha)}
\end{align*}
where we upper bound KL divergence by $\chi^2$ divergence and restrict $\E$ to $\left[0,\frac{3(1-\alpha)}{4}\right]$ (our proposed $\E$ below doesn't exceed $3(1-\alpha)/4$ as $n \geq \log K/(1-\alpha)$). Therefore, we have that
\begin{align*}
\max_{i \in [K]}\mathbb{E}_i \left[ \mathbb{E}_A \hat{L}_T - L_{T,i} \right] \geq \mathcal{E}T \left(1 - \max \left\{ \frac{e}{1+e}, \frac{5n \mathcal{E}^2}{\log (K-1)\alpha(1-\alpha)} \right\} \right)
\end{align*}
Choosing $\E = \sqrt{\frac{e \alpha (1-\alpha)\log (K-1)}{5n(1+ e)}}$ reveals the claimed bound.  
\end{pf}


\begin{lemma}[Lemma 3 from \citet{nicololep}]\label{lem:intermediate}
For any randomized strategy, there exists $D$ deterministic strategies and a probability vector $\beta = (\beta_1,\dots,\beta_D)$ such that for every $t$ and every possible outcome sequence $\{\ell_s\}_{s=1}^{t-1}$, 

\[ \PP_A \left[i_t =i|\{\ell_s\}_{s=1}^{t-1}\right] = \sum_{d=1}^D \beta_d \mathbbm{1}_{\left[i_t^d =i| \{\ell_s\}_{s=1}^{t-1}\right]} \]

\end{lemma}


\begin{pf}[Proof of Theorem \ref{thmlb:lep}]
We will begin by applying the above Lemma \ref{lemlb:lep} with $\nicefrac{\alpha}{2}$ and with the constant $c = 0.36$ (out of convenience) which is indeed lesser than the one we have proven the above lemma for. Note that there is some $j \in [K]$, for which 

\begin{equation}
\mathbb{E}_j [R_T] \geq 0.36 \sqrt{\frac{\alpha}{2}(1-\frac{\alpha}{2})T\log K \frac{T}{n}}\geq 0.09\sqrt{7 \alpha T \log K \frac{T}{n}} \text{ (as } \alpha \leq 1/4) \label{eqn:local}
\end{equation}
 
\noindent
We will now show that under $\PP_j$, the probability that $Q \geq \alpha TK$ is less than $\frac{9}{100T}$. Recall that $\mu_T = \frac{1}{T}\sum_{t=1}^T \ell_t$ and
$Q = \sum_{t=1}^T \norm{\ell_t - \mu_T}_2^2 = \sum_{i=1}^K v_{\alpha,i}$ where $v_{\alpha,i} = \sum_{t=1}^T (\ell_{t,i} - \mu_{T,i})^2$. Noting that $\ell_{t,i} \in \{0,1\}$, we have $v_{\alpha,i} = T \mu_{T,i} (1-\mu_{T,i}) \leq T \mu_{T,i} = \sum_{t=1}^T \ell_{t,i}$. Applying Bernstein's inequality (refer to Theorem 2.10 in \citet{boucheronci} with $b=1,~ v = T(\alpha/2)(1-\alpha/2),~ c = b/3 = 1/3$) along with a union bound gives us that for all $\delta \in (0,1)$, under $\PP_j$, with probability at least $1-\delta$, we have:
\begin{align}
\sum_{t=1}^T \ell_{t,i} & \leq T \left(\frac{\alpha}{2} + \epsilon\right) + \sqrt{2 T \left(\frac{\alpha}{2} + \epsilon\right)  \log \frac{K}{\delta}} + \frac{1}{3} \log \frac{K}{\delta}~. \label{aqw}
\end{align}
for all $i \in \{1,\ldots,K\}$. Now note that by definition of $\epsilon = 0.36 \sqrt{(\alpha/2) (1-\alpha/2) \log (K-1)/n}$ and by the assumption $n \geq 8\log (K-1)/\alpha$,
\[
\frac{\alpha}{2}+\epsilon = \frac{\alpha}{2} + 0.36 \sqrt{\frac{\alpha}{2}\left(1-\frac{\alpha}{2}\right)\frac{\log (K-1)}{n}} \leq 0.59\alpha
\]
Substituting this in \eqref{aqw} above, we get:
\begin{align}
\sum_{t=1}^T \ell_{t,i} &\leq 0.59T\alpha +  \sqrt{2T(0.59\alpha)\log \frac{K}{\delta}} + \frac{1}{3}\log \frac{K}{\delta} \label{asd}
\end{align}
Now we claim that $T\alpha \geq 16 \log \frac{K}{\delta}$ holds for $\delta = \frac{9}{100T}$. This follows from our assumptions that $\alpha \geq \frac{32 \log T}{T}$ and $T \geq \frac{100}{9} K$~. Substituting this back into \eqref{asd}, we can see that $\sum_{t=1}^T \ell_{t,i} \leq T\alpha$~. Hence, this gives us that $Q \leq \alpha TK$.\\
\noindent
Now we will show that there exists a sequence of losses with $Q \leq \alpha TK$ and $\mathbb{E}[R_T] \geq 0.045\sqrt{7 \alpha T \log K \frac{T}{n}}$ where the expectation is taken with respect to the internal randomisation of the strategy. Suppose this were not true, then we would have that $ \indicator{\{Q \leq \alpha T K\}}\mathbb{E}_{j}[R_T | \{\ell_t\}_{t=1}^T] \leq 0.045\sqrt{7 \alpha T \log K \frac{T}{n}}$ (since $\PP_j$ is independent of the internal randomisation). Then we would consequently have:

\begin{align*}
\Ex_{j}[R_T] & =  \Ex_{j}\left[R_T \indicator{\{Q \leq \alpha T K\}} \right] + \Ex_{j}\left[R_T \indicator{\{Q > \alpha T K\}} \right] \nonumber \\
& \leq 0.045\sqrt{7 \alpha T \log K \frac{T}{n}} + T \cdot  \PP_j\!\left( Q > \alpha T K\right)  \nonumber \\
& \leq 0.045\sqrt{7 \alpha T \log K \frac{T}{n}} + 0.09 < 0.09\sqrt{7 \alpha T \log K \frac{T}{n}}
\end{align*}
which contradicts equation \eqref{eqn:local}. Hence, $\Ex[R_T] \geq 0.09\sqrt{7 \alpha T \log K \frac{T}{n}} $.

\end{pf}

\begin{pf}[Proof of Lemma \ref{lemlb:leb}]
As mentioned in the main text, the key difference here from standard bandit lower bounds is that $\sum_{i \in [K]} N_i(t-1)$ (the sum of all revealed labels till time $t-1$) is upper bounded by $n$. Barring this, the proof follows almost identically as that done in \citet{gerchinlatt} but we mention it here for completeness. Consider the following $K$ probability distributions used to construct the stochastic losses. For $i \in [K]$, let $\Q_i$ be a distributions such that under $\Q_i$, $\ell_{t,i}$ is drawn Bernoulli $(\alpha)$ for all $t \in \{1,2,\dots, T\}$, and $\ell_{t,j}$ is drawn Bernoulli $(\alpha + \E)$  for all $t \in \{1,2,\dots, T\},~ j \in [K],~ j \neq i$ (we specify $\E$ later). Additionally, let $\Q_0$ be the joint distribution under which all $\ell_{t,i}$ are i.i.d Bernoulli $(\alpha + \E)$ random variables for $t \in \{1,2,\dots, T\}$ and $i \in [K]$. Also define $\bar{\Q} = \frac{1}{K}\sum_{i=1}^K \Q_i$, the distribution our losses will finally be drawn from. As before, let $\Ex_i$ denote the expectation taken with respect to $\Q_i$. Under (each) $\Q_i$ we have the following:

\begin{align}
\Ex_i \left[\hat{L}_T - \min_{j \in [K]} L_{T,j}\right] &\geq \Ex_i\!\left[\hat{L}_T \right] - \min_{j \in [K]} \Ex_i\!\left[ L_{T,j}\right] = \Ex_i\!\left[\sum_{t=1}^T \ell_{t,i_t} \right] - \min_{j \in [K]} \Ex_i\!\left[ \sum_{t=1}^T \ell_{t,j} \right] \nonumber \\ 
&= \sum_{t=1}^T \Ex_i\bigl[ \alpha + \E - \E \indicator{\{i_t=i\}} \bigr] - T \alpha    \nonumber \\ 
&= T \E \left( 1 - \frac{1}{T} \sum_{t=1}^T \Q_i(i_t = i) \right)~, \label{eqn:b}
\end{align}
\noindent
Now, we can further lower bound the above expression by appealing to Pinsker's inequality which tells us that $\Q_i (i_t = i) \leq \Q_0 (i_t = i) + (\KL(\Q_0^{i_t}, \Q_i^{i_t})/2)^{1/2}$~~\footnote{$\Q_i^{i_t}$ denotes the probability measure of $i_t$ under $\Q_i$} for all $t \in \{1,2,\dots, T\}$ and all $i \in [K]$. We substitute this in \eqref{eqn:b}, average over $i \in [K]$ in order to bound the regret under $\bar{\Q}$, and use the concavity of the square root to yield:

\begin{equation}
\Ex_{\bar{\Q}}\!\left[\hat{L}_T - \min_{j \in [K]} L_{T,j}\right] \geq T \E \left( 1 - \frac{1}{K} - \sqrt{\frac{1}{2T} \sum_{t=1}^T \frac{1}{K} \sum_{i=1}^K \KL\bigl(\Q_0^{i_t}, \Q_i^{i_t}\bigr)} \, \right) \label{eqn:a}
\end{equation}
\noindent
Now we will upper bound the KL divergence terms:
\begin{align*}
\KL\!\left(\Q_0^{i_t}, \Q_i^{i_t}\right) & \leq \KL \left(\Q_0^{(h_t,i_t)}, \Q_i^{(h_t,i_t)}\right)
= \Ex_{\Q_0} \bigl[N_i(t-1)\bigr] \KL \bigl( \Ber(\alpha+\epsilon), \Ber(\alpha) \bigr) \nonumber \\
& \leq \Ex_{\Q_0} \bigl[N_i(t-1)\bigr] \, \frac{\E^2 }{\alpha(1-\alpha)}~, 
\end{align*}
\noindent
where the first inequality follows from the Data Processing Inequality and second by upper bounding the KL divergence by the $\chi^2$ divergence. $h_t$ denotes the history available at time $t$ and $N_i(t-1)$ refers to the number of pulls of arm $i$ till time $t-1$. We now average the above quantity over $i \in [K]$ and $ t \in \{1,2,\dots, T\}$ to yield:

\[
\frac{1}{T} \sum_{t=1}^T \frac{1}{K} \sum_{j=1}^K \KL\bigl(\Q_0^{i_t}, \Q_i^{i_t}\bigr) \leq \frac{1}{T} \sum_{t=1}^T \frac{n \E^2}{K \alpha(1-\alpha)} \leq \frac{n \E^2}{ K \alpha(1-\alpha)}~.
\]
\noindent
The above equation incorporates the strict restriction on the revealed labels as $\sum_{i \in [K]} N_i(t-1)$ is upper bounded by $n$. Plugging the above inequality into \eqref{eqn:a} and substituting $\E = (1/2\sqrt{2}) \sqrt{\alpha (1-\alpha) K/n}$ gives us the claimed bound.

\end{pf}


\begin{pf}[Proof of Theorem \ref{thmlb:leb}]
The proof follows almost identically as Theorem \ref{thmlb:lep}.
\end{pf}





\end{document}